\documentclass[]{Liebert_Author}

\setcitestyle{numbers}

\usepackage{graphicx} 
\usepackage{caption}
\usepackage{subcaption}
\usepackage{todonotes}
\usepackage{soul,color}
\usepackage{colortbl}
\usepackage{stfloats}

\usepackage{geometry}
\renewcommand\cite{\citep}
\renewcommand\citep[1]{\textsuperscript{\citealp{#1}}}
\usepackage{setspace}
\usepackage{tabularx}
\usepackage{tablefootnote}
\hypersetup{
  colorlinks=true,
  linkcolor=black,
  citecolor=black,
  urlcolor=black,
}

\date{}
\usepackage{titlesec}
\usepackage{setspace}
\usepackage{lipsum}
\titleformat{\section}
  {\normalfont\bfseries} 
  {\thesection} 
  {1em} 
  {} 

\titleformat{\subsection}
  {\normalfont} 
  {\thesubsection} 
  {1em} 
  {} 

\titleformat{\subsubsection}
  {\normalfont\itshape} 
  {\thesubsubsection} 
  {1em} 
  {} 

\titlespacing{\section}
  {0pt} 
  {12pt plus 4pt minus 2pt} 
  {8pt plus 2pt minus 2pt} 

\titlespacing{\subsection}
  {0pt} 
  {12pt plus 4pt minus 2pt} 
  {8pt plus 2pt minus 2pt} 

\titlespacing{\subsubsection}
  {0pt} 
  {12pt plus 4pt minus 2pt} 
  {8pt plus 2pt minus 2pt} 

\color{black}
\title {Towards a Unified Naming Scheme for Thermo-Active Soft Actuators: A Review of Materials, Working Principles, and Applications}  

\author{Trevor Exley,$^{1\ast}$ Emilly Hays,$^{1}$ Daniel Johnson,$^{1}$ Arian Moridani,$^{1}$, Ramya Motati,$^{1}$\\ and Amir Jafari$^{1}$\\
{$^{1}$Advanced Robotic Manipulators (ARM) Lab, the Department of}\\
{Biomedical Engineering, University of North Texas, Texas, United States}\\
{$^\ast$To whom correspondence should be addressed; E-mail:  trevorexley@my.unt.edu.}
    }

\begin{document}
\setstretch{1.0}
\twocolumn[
\begin{@twocolumnfalse}

\maketitle
\thispagestyle{empty}
\pagestyle{empty}
\begin{abstract}
Soft robotics is a rapidly growing field that spans the fields of chemistry, materials science, and engineering. Due to the diverse background of the field, there have been contrasting naming schemes such as 'intelligent', 'smart' and 'adaptive' materials which add vagueness to the broad innovation among literature. Therefore, a clear, functional and descriptive naming scheme is proposed in which a previously vague name --- \textit{Soft Material for Soft Actuators} --- can remain clear and concise --- \textit{Phase-Change Elastomers for Artificial Muscles}. By synthesizing the working principle, material, and application into a naming scheme, the searchability of soft robotics can be enhanced and applied to other fields.  The field of thermo-active soft actuators spans multiple domains and requires added clarity. Thermo-active actuators have potential for a variety of applications spanning virtual reality haptics to assistive devices. This review offers a comprehensive guide to selecting the type of thermo-active actuator when one has an application in mind. Additionally, it discusses future directions and improvements that are necessary for implementation.

\end{abstract}
\end{@twocolumnfalse}
]



\maketitle

\section{Introduction}\label{Intro}

\begin{figure*}[b]
	\centering
		\includegraphics[width=.95\textwidth]{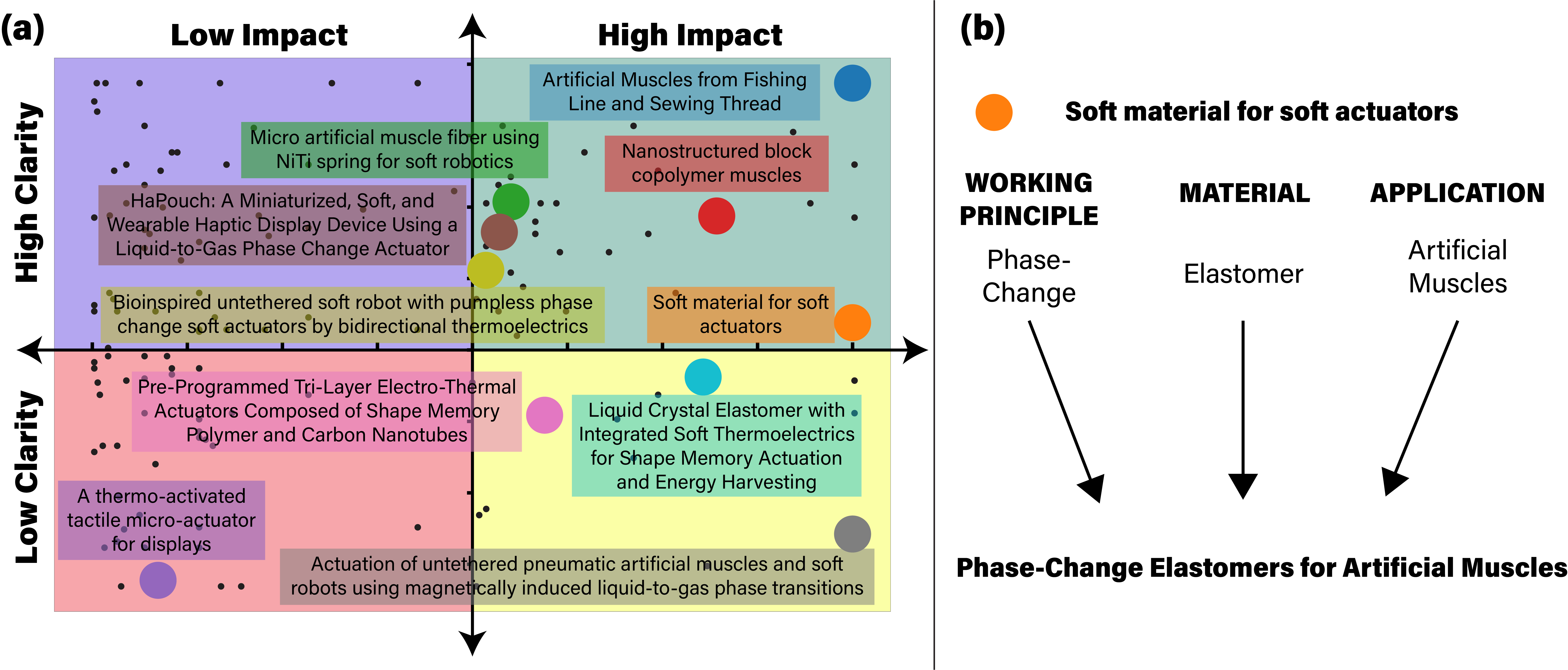}
    	  \caption{(a) Quadrant chart comparing published papers based on naming scheme in terms of clarity and impactfulness, black dots represent the 130 papers scraped, with select ten titles with larger colored dots labeled. (b) Sample implementation of proposed naming scheme.}
   \label{quadrant}
\end{figure*}
There has been an intense interest in soft robotics and soft actuators since the turn of the 21st century; however, it is difficult to navigate the literature surrounding these topics, as the terminology related to soft actuators is constantly redefined and interchanged. This can lead to confusion for those searching for related papers, unsure of which keywords are most used in the field, or cause uncertainty about the definitions of terms that have not been clearly defined. In the field of soft robotics, it is common that the naming schema consists of the stimulus (e.g., \textit{pneumatic, hydraulic, electrothermal, electromagnetic,} etc.) followed by 'soft actuator' or 'soft robot'. An equally popular case is using the main material as the preceding term (e.g., \textit{dielectric elastomer, shape-memory polymer, liquid-crystal elastomer, phase-change,} etc.). More importantly, we can target 'soft robotics' as a field and see that 'smart materials', has been almost made synonymous with words like 'intelligent materials', 'responsive materials', and 'adaptive materials'.  This interchangeability of smart, intelligent, and responsive may not lead to confusion from the reader, but causes fragmentation in the literature. For example, when conducting an electronic search for 'responsive materials; soft robotics', you may find a plethora of journals primarily focused on chemistry rather than soft robotics \cite{LEE2020100258,Shen_Chen_Zhu_Yong_Gu_2020,Tognato_Armiento_Bonfrate_Levato_Malda_Alini_Eglin_Giancane_Serra_2018}, whereas if one searches 'intelligent materials; soft robotics', more robotics-centric journals appear in the search results \cite{6595683,Zhao_2021jrm,KHALID2022113670}. This is not to imply there is not a prominent overlap in the fields, but this segmentation can make it more difficult to find relevant work if there is not uniformity in titles and keywords. Although new soft actuator technology is constantly developing, a persistent issue lies in the inconsistency of naming conventions. This issue not only confuses readers but also forms a barrier that can prevent researchers from readily accessing relevant work in the field.


This naming inconsistency is a complex debate, as clarity and impact of a paper are not always correlated. Previous works have investigated the effects on paper quality and external factors on the number of citations a study receives, as well as the number of accesses, downloads, or the overall 'use' of a paper \cite{Tahamtan_Safipour_Afshar_Ahamdzadeh_2016}. For example, the title of a paper has unique effects in relation to overall impact (number of citations), such as the length negatively affecting interactions, and combinatory titles (i.e., use of separator such as a hyphen or colon separating ideas) positively affecting citations  \cite{Rostami_Mohammadpoorasl_Hajizadeh_2013,jamali2011article}.  Furthermore, and perhaps even more significant, is ones choice of associated keywords. For example, Jacques et al. states that having two keywords other than the words included in the title drastically increases findability and, therefore, citations of a paper especially with electronic searches being the norm \cite{jacques2010impact}. Therefore, to maximize clarity, careful keyword selection should be an indispensable part of the publication process in soft robotics research.
The overall 'impact' of an article is very important when it comes to the number of citations it might receive. Impact can consist of many different things, such as author fame/achievement \cite{RePEc:spr:scient:v:69:y:2006:i:3:d:10.1007_s11192-006-0166-1,https://doi.org/10.1002/asi.23209}, use of 'attention grabbers', or marketability/branding of a technology \cite{doi:10.1509/jmkg.71.3.171}.

A title should be clear and impactful to avoid confusion and be memorable within the literature. Figure \ref{quadrant} depicts the spread of papers within the field within a chart that is divided into quadrants representing examples that are (I) clear and impactful, (II) clear but not impactful, (III) unclear and not impactful, and (IV) unclear but impactful. The metrics for categorizing the papers among the different quadrants are specified in section \ref{namingscheme}.

For a soft actuator to be considered 'thermo-active', the primary actuation principle should be heat. In this sense, if the final goal of a device is actuation, the penultimate domain should be thermal (regardless if previous domain is electrical), i.e., the heating source can be swapped without critical adjustment of geometry or properties. Additionally, the majority of the actuation structure should be soft for consideration of this review. This should not be confused with electrothermal soft actuators, as they would be considered a specific subtype of thermo-active actuators \cite{act8040069}. Thermo-active soft actuators are significant as they are able to be made completely soft (including heating stimuli), require relatively low input power ($< 10W$), and therefore can be made portable for applications.

Soft robotics have been targeted as a sect of mechanics that can solve problems in physical human-robot interactions (pHRI) such as prosthetics, drug delivery, and rehabilitation. To be classified as 'soft', materials should have a Young's modulus comparable to soft biological tissue (skin, cartilage, muscles, etc; $\leq$ 1GPa) \cite{Rus2015-vy}. In the areas of prosthetics and rehabilitation, the performance of the thermo-active soft actuators should resemble human skeletal muscles\cite{Alexander1977-ij} (energy density $\geq10\frac{kJ}{kg}$, bandwidth $\geq$ 10Hz, and efficiency $\geq$ 10\% ).

Breakthroughs in material fabrication and manufacturing processes have led to achieving more complex geometries and advanced control systems inherent to the material itself.  The development of functional materials simultaneously simplifies and complicates actuation mechanisms by decreasing the number of elements required for fabrication, but increasing many considerations at the design level. These advancements further underline the need for a standardized naming convention, which considers not just the material but also the actuation mechanisms and design considerations.

Community consensus and adoption of a naming scheme is a lengthy process and cannot be achieved overnight. Therefore, it is crucial that young researchers have tools to traverse current literature and find areas needing improvement and understand what technologies are suitable for certain applications. The body of this review will go over current technologies for thermo-active soft actuators, with focus on materials and working principles.


\section{Review Structure}

This review strives to accomplish three tasks. \textbf{(I)} Develop a universal naming scheme for classifying newly-developed thermo-active soft actuators (any soft actuator controlled by heating). \textbf{(II)} Present an application-based metrics evaluation for readers to understand which technology can accomplish a specific task. \textbf{(III)} Discuss strengths and drawbacks to each thermal technology, and highlight future trends. 
\begin{enumerate}[label=\textbf{\roman*},itemsep=0pt,parsep=0pt,topsep=0pt,partopsep=0pt]
    \item \textbf{UNIVERSAL NAMING: Working principle, material and application of a soft actuator should be clarified in the functional name of an actuator.} Thus, naming of such devices can be consistent, informative, and have broad applicability across literature.
    \item \textbf{TECHNOLOGY EVALUATION: Applications of soft actuators are a critical consideration for the selection of different technologies.} The strengths and drawbacks of thermo-active soft actuators span many metrics making technology suitable for delicate manipulation (stiffness, force) as well as adaptive locomotion (bandwidth, portability).
    \item \textbf{POTENTIAL AND LIMITATIONS: Thermo-active actuators have great potential as they can be made fully portable, and active cooling has not been fully characterized.} Energy can be transferred into these actuators through a variety of methods such as through light, magnetism, and this technology should be fully realized even in extreme environments (cold or hot).
\end{enumerate}

It is important to note the inclusion and exclusion criteria of this review, light-activated actuators and small-scale applications \cite{https://doi.org/10.1002/adma.201603483} are excluded, but are further discussed when it comes to unique materials that are developed. An overview of the different thermo-active technologies are discussed and future trends are discussed. 

\subsection{Naming Scheme}\label{namingscheme}
First, the standards that make a good naming scheme should be overlaid. A good name should be (i) descriptive and informative, (ii) consistent, (iii) clear and simple, (iv) broadly applicable, (v) flexible, and (vi) share a community consensus \cite{https://doi.org/10.1111/avsc.12179}. However, these general guidelines can be challenging to consolidate for developing a new technology. Therefore, it is necessary to further simplify these standards for thermo-active soft actuators. There are many features of thermo-active soft actuators that are critical for understanding the mechanics behind actuation such as materials, working principle, actuator type, heating stimuli, and application. The primary areas of focus for development have been (I) materials, (II) working principle, and (III) application. Thus, Eq. \ref{eq:naming} can be developed which overlays the general structure a name should have:

\begin{equation}
    [\text{Working Principle}] + [\text{Material}] + [\text{Application}]
    \label{eq:naming}
\end{equation}

However, it is understandable that studies may not always innovate on all three factors, or have one of these factors as a control where different aspects are controlled and tested for. Therefore, for a paper to be considered clear it must satisfy $\frac{2}{3}$ of the requirement of Eq. \ref{eq:naming}. Impactfulness, on the other hand is a more difficult metric to quantify. For a journal to be considered impactful it should equal or surpass the respective journal's impact factor.


In a rapidly growing field, it is important to not only have clarity in the description of a technology, but also be clear in the novelty of the invention. Thus, for select studies, novelty will be presented and there will be discussion on the overlap between these three naming criteria. Currently the best way to search for a specific thermo-active soft actuator is to search by either material or working principle, but this is a problem, as it may be unclear to young researchers what options there are. Therefore, a preliminary literature search is conducted to provide a starting point for navigating current literature.

\subsection{Literature Search}

\begin{figure}[h]
    \centering
    \includegraphics[width=.45\textwidth]{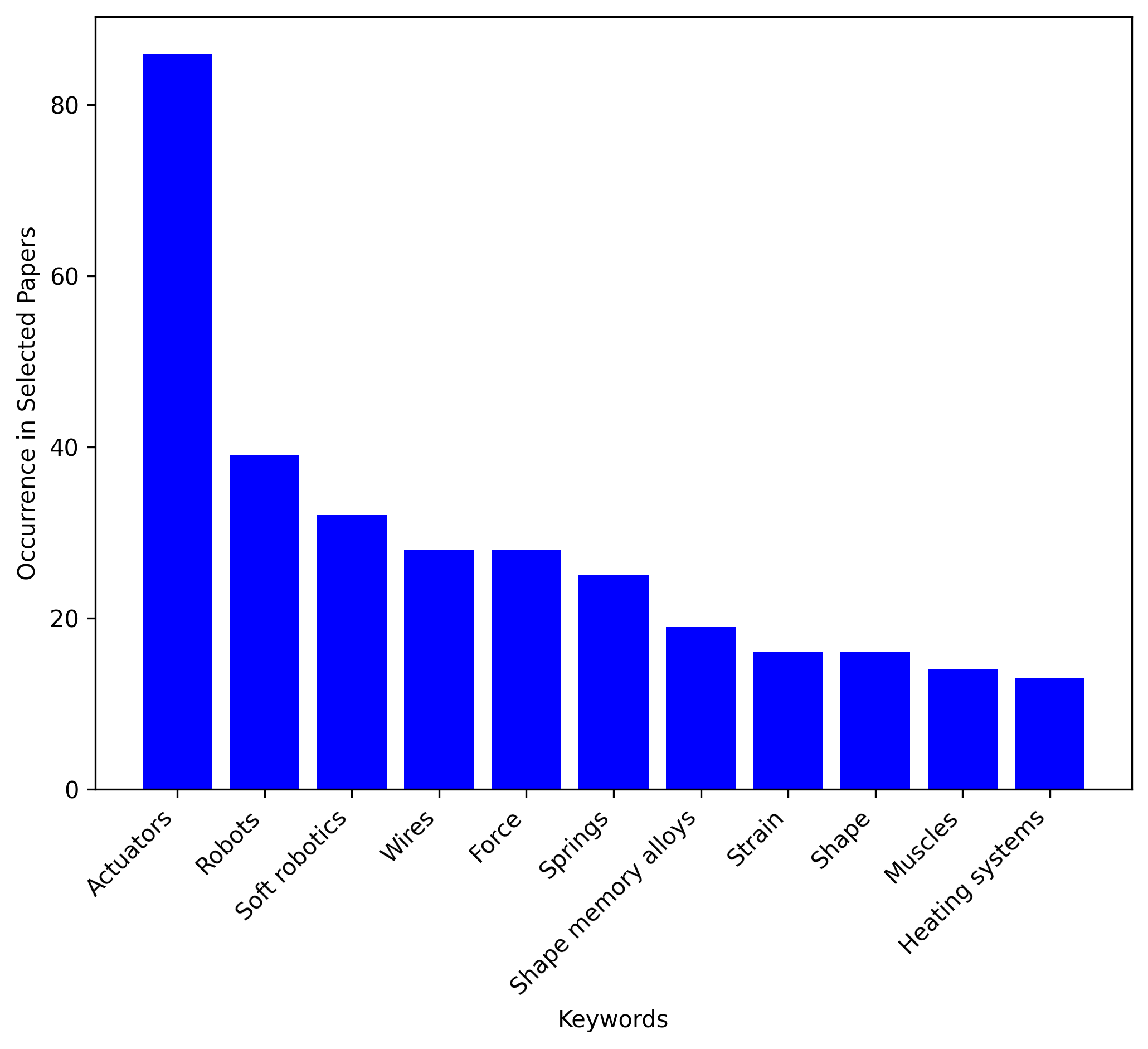}
    \caption{Keyword Frequency in 130 Papers on Thermo-Active Robotics}
    \label{fig:enter-label}
\end{figure}

Using the IEEE Xplore Metadata Search API, search engines were scraped using predetermined search queries. The selected queries include "shape memory alloys in soft actuators", "thermally responsive hydrogels in actuators", "liquid crystal elastomers and thermal actuation", and "phase change fluid in actuators". These texts were chosen to provide an overview of thermo-active technology and serve as an example of queries young researchers would choose, and the subsequent results can be replicated by following similar methodology. This scraping is to quantify the clarity, impact, and frequency of keywords among relevant literature. The Flesch Reading Ease scoring \cite{fleschNewReadabilityYardstick1948} (Eq. \ref{eq:flesch}) is a basic quantification of clarity of the title, taking into consideration the total syllables and words of the title. For impact, the number of citations is divided by the number of years since publication. Both clarity and impact scores are winsorized at the 5th and 95th percentiles, scaled using a standard scaler, and rescaled along the impact axis to show spread within the quadrants. Additionally, 10 papers were selected for further discussion of our proposed naming scheme, which can be found in Section \ref{discussion}. All of this information was used to construct the quadrant chart featured in Fig. \ref{quadrant}. We also obtained index terms were obtained from the API for each paper to rank the frequency of keywords across multiple papers (Fig. \ref{fig:enter-label}). These terms include both author-defined keywords, and IEEE terms ("keywords assigned to IEEE journal articles and conference papers from a controlled vocabulary created by the IEEE"). The maximum records scraped for each term was 200.

\begin{equation}
    \label{eq:flesch}
    206.835 - 1.015(\frac{\text{total words}}{\text{total sentences}}) - 84.6(\frac{\text{total syllables}}{\text{total words}})
\end{equation}

\section{Materials and Working Principles}
Of the three criteria, the materials and working principle provide more clarity than application, as there is significant overlap of the specific applications (e.g., liquid crystal elastomer bending actuator, shape memory bending actuator).
\subsection{Materials}
This section delves into the materials utilized in thermo-active robotics, shedding light on the underlying concepts of shape memory, working principles and phase change behavior. It explores the environmental stimuli of shape memory polymers, transition temperatures and their role in alloys memory, and types of fluid selections for phase change actuation applications. The selected variations of thermo-active soft actuators are compiled to Table \ref{tbl:applications}.
\subsubsection{Shape Memory}
\begin{figure}[h]
    \centering
    \includegraphics[width=.45\textwidth]{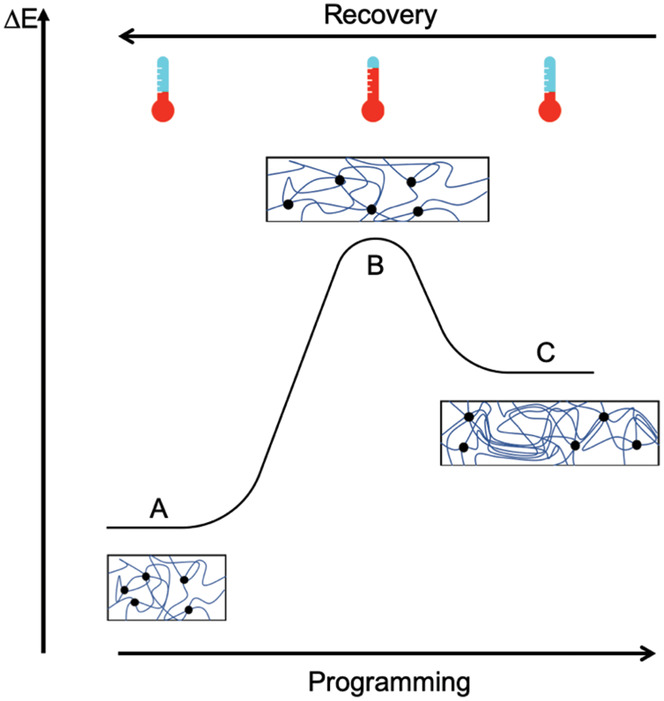}
    \caption{Strain graph showing the recoverable deformation of shape memory polymers. Reproduced from Delaey et al.\cite{https://doi.org/10.1002/adfm.201909047} with permissions.}
    \label{fig:shapememory}
\end{figure}

Shape memory polymers are similar to shape memory alloys in that their form is dictated by the environment they are subjected to, typically the stimulus that institutes change is heat. These materials rely on transition temperatures which mark configurations within the range (Fig. \ref{fig:shapememory}). In addition heat, pH, elasticity, magnetics, light, humidity, and redox reactions can cause configuration changes to the polymers. Furthermore, these polymers can be used to house a mechanical working principle through various structural configurations, enabling actuation via a different mechanism i.e phase change of a fluid within a polymer container. Two-way shape memory \cite{ZARE2019706} works with applying an indirect or direct stimuli such as heat, light, or chemical materials, to produce an output reaction. The external stimulus, heating for example, triggers a change in the permanent shape to a temporary one. Conversely, applying the reverse stimulus, cooling, causes the temporary shape to go back to the permanent shapes memory \cite{ZARE2019706}. 
Polyester Urethane \cite{https://doi.org/10.1002/macp.201200096}, is biocompatible with the human body and flexible, making it suitable for a wide range of biomedical applications.
Reversibility is important \cite{https://doi.org/10.1002/adma.201300880} and critical in applications such as vascular grafts or artificial heart valves to have the ability to repetitively change and recover its shape.
Additive materials such as CNTs have been implemented with SMP technology \cite{doi:10.1089/soro.2018.0159,10.1063/1.5033487} to enhance mechanical properties, and improve thermal and electrical conductivity within the system. The synergy between CNTs and SMPs open possibilities for smart implants, drug delivery systems and tissue scaffolds with shape memory properties.

\subsubsection{Block Copolymers}
\begin{figure}[h]
    \centering
    \includegraphics[width=.45\textwidth]{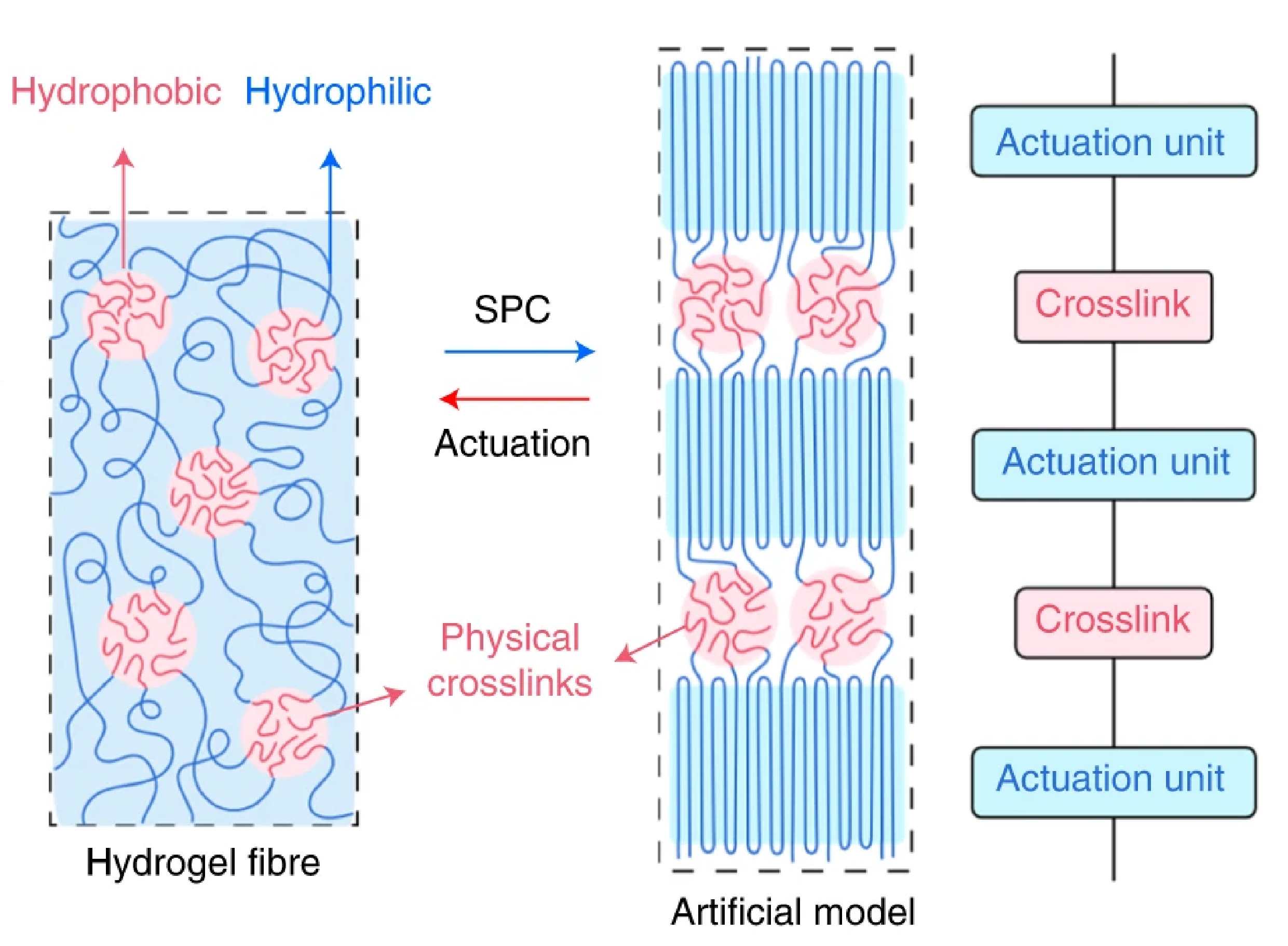}
    \caption{Block copolymer fiber characterized by hydrophobic and hydrophilic regions for reversible strain-programmed crystallization (SPC) and actuation. Reproduced from Lang et al.\cite{Lang2022-jy} with permission.}
    \label{fig:copolymer}
\end{figure}

Block copolymers that resemble artificial muscles \cite{Lang2022-jy,Lin2022} typically consist of two or more chemically distinct polymer blocks that are covalently bonded together in single chains (Fig. \ref{fig:copolymer}). These block consists of various chemical compositions, weights and physical properties. Exhibiting muscle-like behaviors with copolymers, the designed and synthesized to contract and extend, changing shape in a controlled manner.  Challenges with this application is calibrating the responsiveness of the apparatus, durability and scaling for mass replication production.

\subsubsection{Liquid Crystal Elastomer}
\begin{figure}[h]
    \centering
    \includegraphics[width=.45\textwidth]{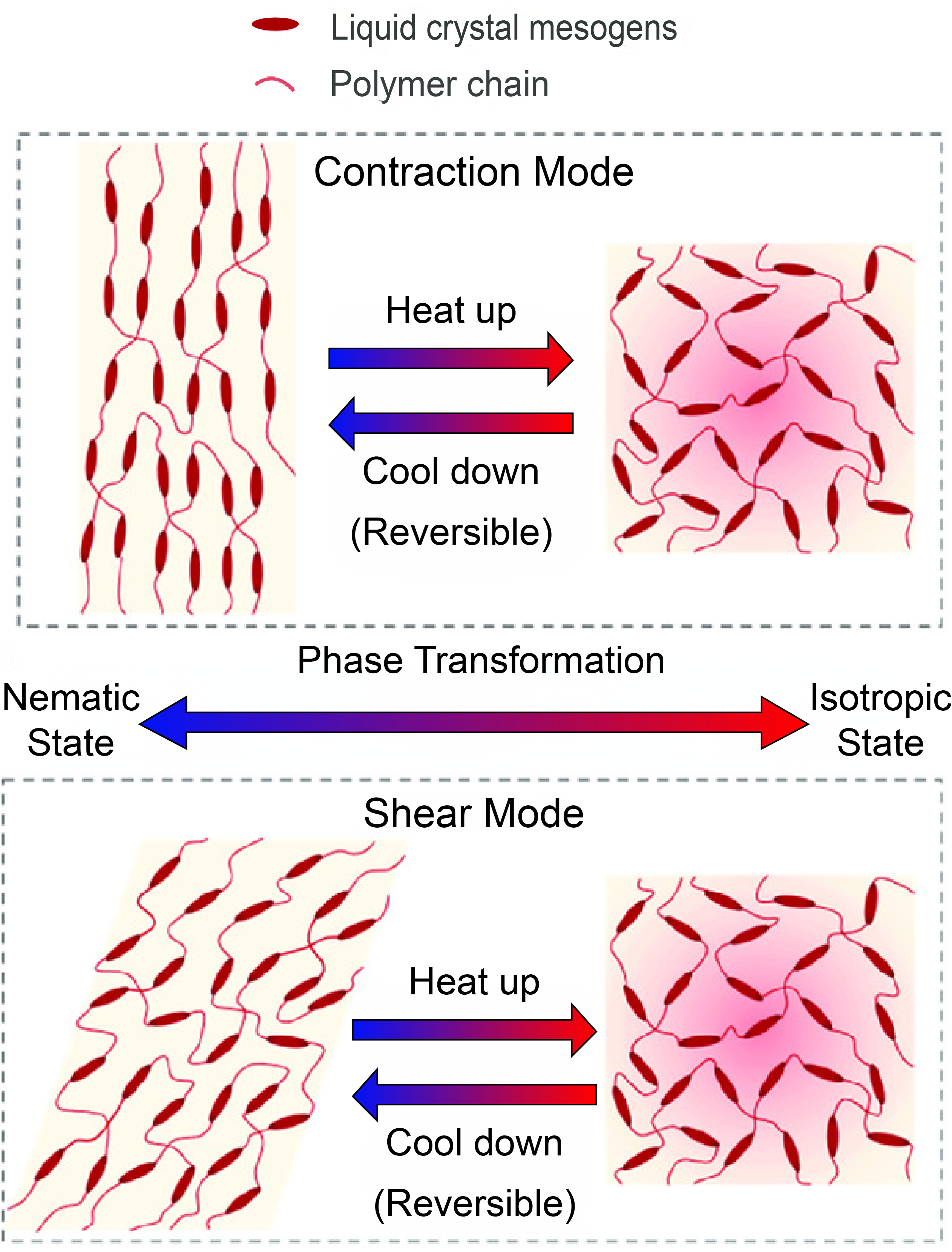}
    \caption{Programmable actuation of liquid crystal elastomers between contraction and shear actuation types based on deformation and heating. Reproduced from Wang et al.\cite{https://doi.org/10.1002/aisy.201900177} with permission.}
    \label{fig:lce}
\end{figure}

Liquid crystal elastomers (LCEs) are composed of long, chain-like molecules aligned through mechanical or thermal processing. The liquid crystal segments within the structure change the structures orientation in response to external stimuli, such as heat or light (Fig. \ref{fig:lce}). The reversible shape allows for the structure to bend and contort for various applications such as soft robotics or artificial muscles.

\subsubsection{Thermally-Responsive Hydrogels}
\begin{figure}[h]
    \centering
    \includegraphics[width=.45\textwidth]{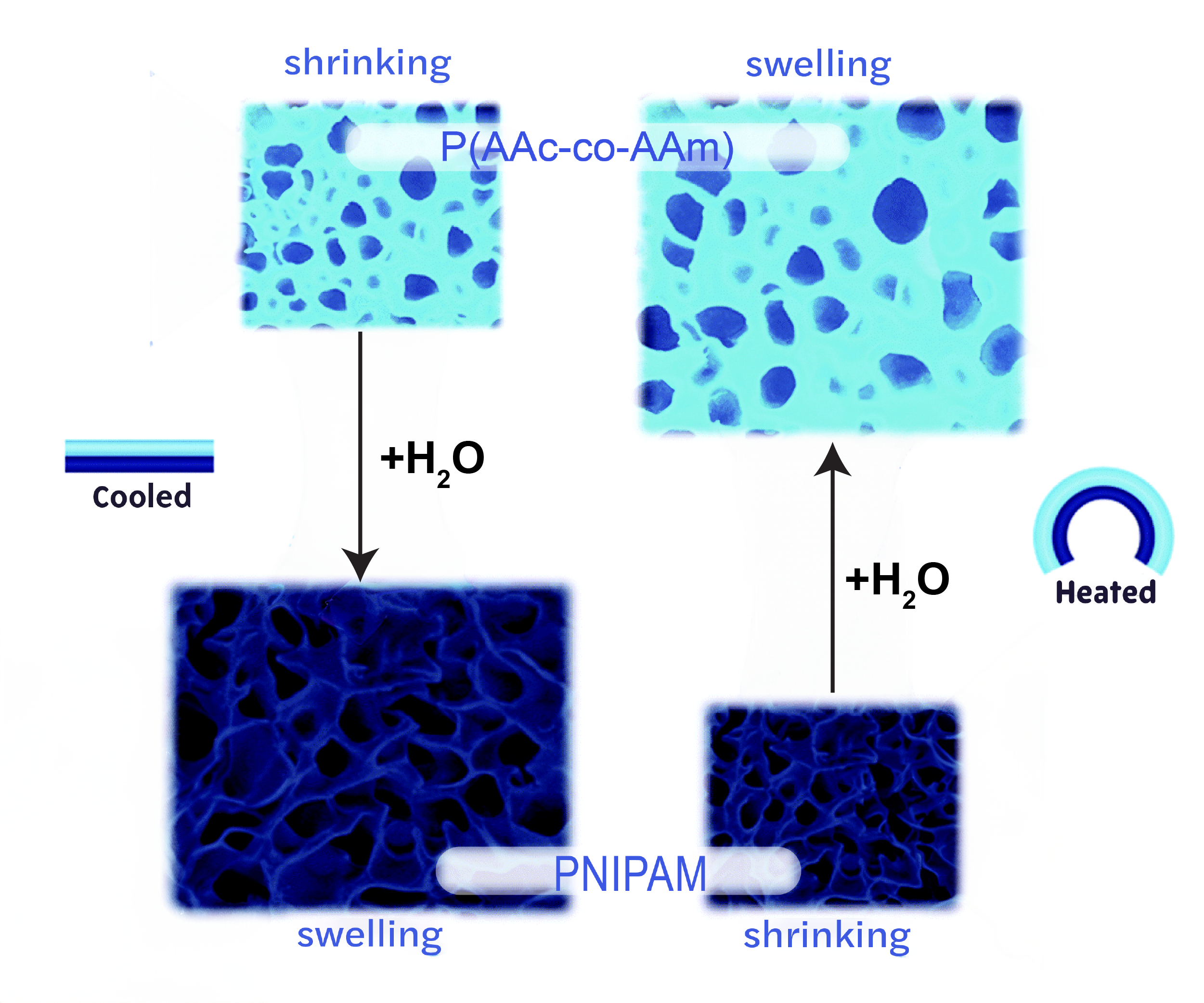}
    \caption{Reversible, antagonistic design utilizing thermo-responsive hydrogels depicting swelling at low temperatures and shrinking at high temperatures for active layer.  Reproduced by Zheng et al.\cite{zheng2018mimosa} with permission.}
    \label{fig:hydrogel}
\end{figure}

Thermally-responsive hydrogels are smart materials that are capable of physical property changes, such as swelling and collapsing in response to a change in temperature. The hydrogels are fabricated with a specific lower (LCST) or upper (UCST) critical solution temperature \cite{zheng2018mimosa} (Fig. \ref{fig:hydrogel}). For the LCST hydrogels, above the temperature causes shrinkage and colder temperatures cause swelling. For UCST hydrogels, above the temperature does the opposite and causes the hydrogel to swell and absorb water, and below that value they shrink and dehydrate\cite{LEE2020100258}. By exploiting the ability to respond to various temperatures, this material can be used for various applications like precise control over microfluidics, artificial muscles, and drug delivery. 

These actuating hydrogels can achieve complex deformations, both with contraction/expansion and bending. The primary limitation of hydrogels are their requirement of an aqueous solution for the actuation cycles to be repeatable.

\subsection{Working Principles}

Shape Memory Alloys are intelligent materials that once deformed can be converted back into their original configuration after exposure to an external stimulus, which is most often heat.
This conversion typically relies on transition temperatures that mark configurations between the ranges made by each transition temperature value. The number of transition temperatures the material has dictates how many forms the material can hold\cite{act9010010}. 

Note that Lang et al.\cite{Lang2022-jy} has multiple stimuli among the working principles, including heat, hydration, and water vapor. This is a unique solution to achieving better actuation capabilities, leaving interpretation for if it is considered to be 'thermo-active'.

\subsubsection{Phase-Change}
 It is difficult to label each part of the research. The elastomer shape\cite{decroly2021optimization}, size\cite{Uramune2022-iv}, and composition\cite{noguchi2020soft, Yoon2023-jy, tang2022wireless} all can affect the final product's actuation. The phase change fluid can also affect the final actuation by changing the fluid used\cite{decroly2021optimization}, and the volume of fluid used\cite{mirvakili2020actuation, Usui_2021}.

 \begin{figure}[h]
	\centering
		\includegraphics[width=.45\textwidth]{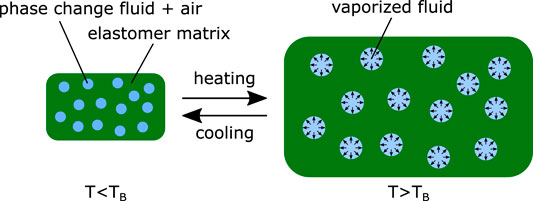}
	  \caption{Working principle of phase-change elastomer acutators showing elastomer-fluid composite below and above the vaporization temperature. Reproduced from Decroly et al.\cite{decroly2021optimization} with permission.}
   \label{phase-change}
\end{figure}

 For a given phase change actuator, the phase change aspect is only looked at in the form of a volumetric expansion in response to thermal stimulus (Fig. \ref{phase-change}). While the fluid used can be viewed as part of the materials used, the working principle here is clearly the volumetric expansion from the fluid to vapor transition. For the proposed naming scheme of [working principle] + [material] + [application], because phase change fluid is the working principle and part of the material, but no other aspect would properly describe the working principle accurately, phase change fluid would still be listed as the working principle and elastomer as the material. There would be no need to mention the fluid for both the working principle and the material, even if it could be viewed as both.

Lets take a closer look at working principle. This would describe how a given actuator actuates. Traditional phase change actuators work by volumetric expansion. For many studies this refers to the fluid used, and the fluids used are typically chosen based on boiling point\cite{decroly2021optimization,8404934}, molecular structure\cite{hiraki2020laser}, vapor pressure\cite{matsuoka2016development}, and toxicity. All these properties are chosen based on the fluid transitioning from a fluid to a vapor, which is the working principle.
Another way to look at this would be that the material’s purpose is to control the volumetric expansion, the fluid’s purpose is to cause the volumetric expansion, and the applied heat is used to initiate the volumetric expansion\cite{doi:10.1089/soro.2020.0018}. Where the working principle would be a combination of the applied heat and liquid to vapor phase change.

It is important to note that although this is within the phase-change subsection, many material developments are being produced \cite{https://doi.org/10.1002/adem.202100863} which include variations of the phase-change principle, not only going between liquid and vapor, but also waxes going from solid to liquid to vary overall stiffness \cite{8288847,doi:10.1089/soro.2020.0080}.

\subsubsection{Electrothermal}

\begin{figure}[h]
    \centering
    \includegraphics[width=0.45\textwidth]{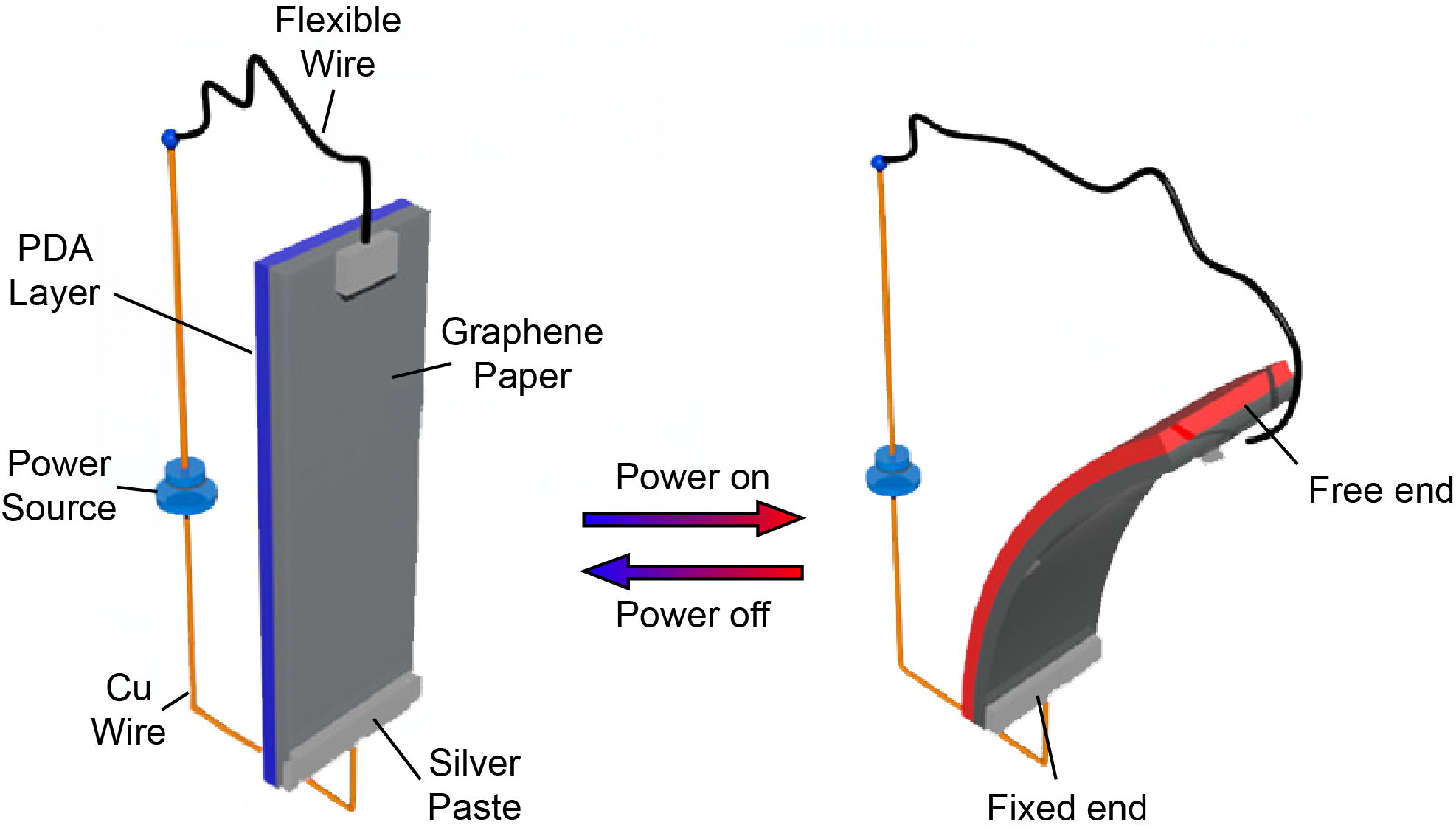}
    \caption{Showing principle of electrothermal film bending actuator utilizing graphene paper as a strain-limiting layer. Reproduced from Liang et al.\cite{doi:10.1021/nn3006812} with permission.}
    \label{fig:electrothermal}
\end{figure}

Electrothermal actuators are one type of soft actuators that produce relatively quick responses, large strains, and are easy to manufacture when compared to other types of soft actuators\cite{Tian2021-hi}. This ease of manufacturing also includes the incorporation of strain-limiting layers and embedded sensors (Fig. \ref{fig:electrothermal}). Electrothermal actuators are mainly controlled through Joule heating.

\section{Applications of Thermo-Active Actuators}
In the field of soft robotics, applications of rehabilitation and assistive devices are popular, as the inherent compliance and adaptability is very beneficial for physical Human-Robot Interactions (\textit{pHRI}). Thus, the characteristics of interest for these devices typically resemble human skeletal muscles, with main metrics being energy density (40 $\frac{J}{kg}$ \cite{Rich2018-hr}), deformation (20-40$\%$ \cite{doi:10.1126/science.aaw2502}), bandwidth (1-10$Hz$ \cite{Rich2018-hr}), power (50-300 $\frac{kW}{m^{-3}}$ \cite{Rich2018-hr}), and efficiency (20$\%$ \cite{Rich2018-hr}). The other unique feature of skeletal muscles is that they can only achieve contraction, and must synergize with agonist-antagonist designs to accomplish complex motion. In this capacity, these metrics should also include things such as portability, safety, and size. Thus, artificial muscles \cite{https://doi.org/10.1002/adma.201704407} should be bio-inspired.

\begin{figure}[h]
    \centering
    \includegraphics[width=0.45\textwidth]{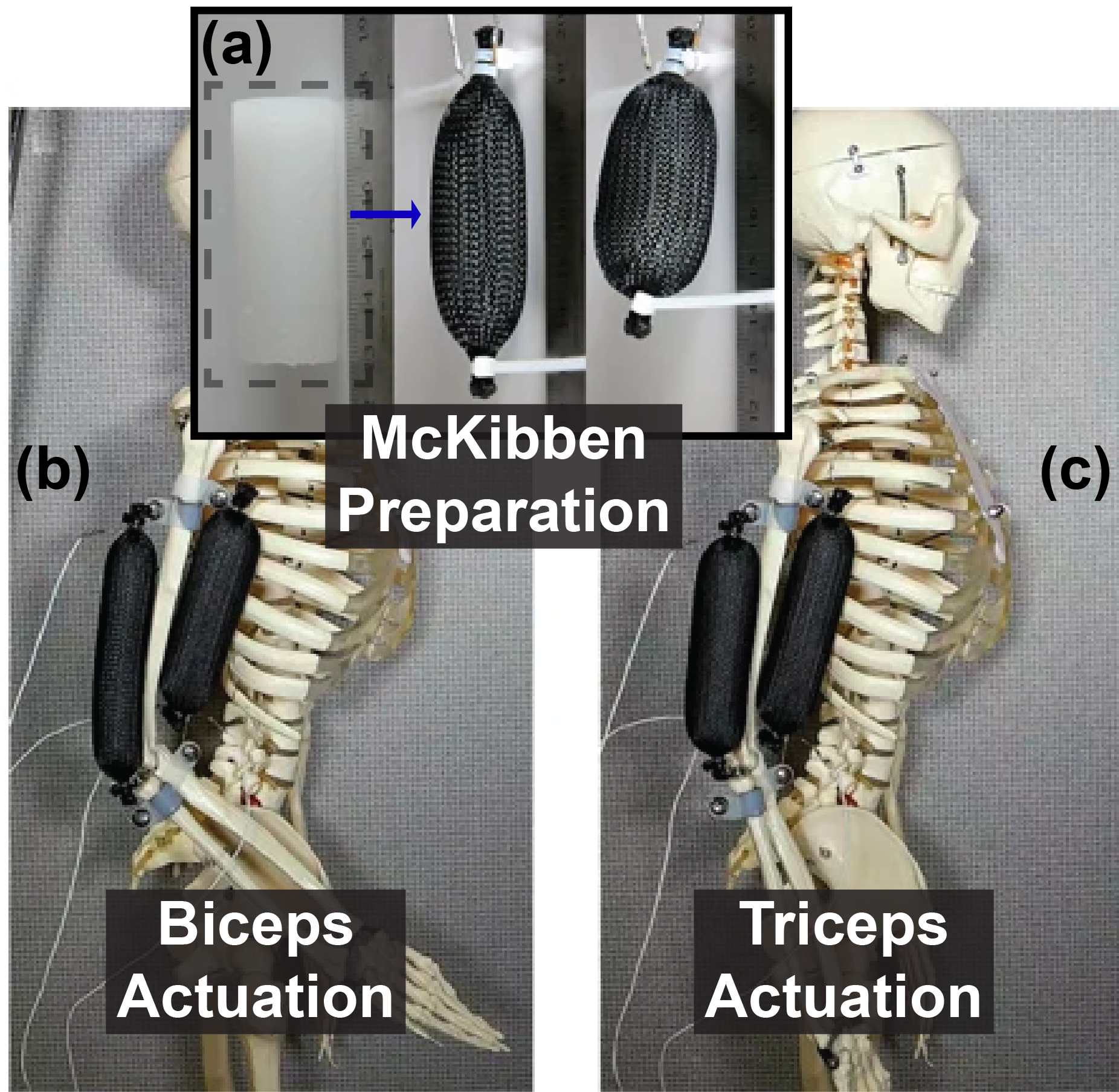}
    \caption{(a)Phase-change elastomer composite enclosed in a braided mesh and implemented into a antagonist-agonist design showing (b) biceps actuation for flexion and (c) triceps actuation for extension.  Reproduced from Miriyev et al.\cite{Miriyev2017-ps} with permission.}
    \label{fig:sm4sa}
\end{figure}

Phase-change \cite{10.1115/FPMC2016-1702} has been incorporated to create artificial muscles (Fig. \ref{fig:sm4sa}). Although many artificial muscles are characterized by a McKibben braided mesh, various bending actuators have been developed by introducing strain limiting layers, such as \cite{8351924} which implements SMA.

There are some assumptions that can be made when it comes to an application. For example, if a work claims to be for rehabilitation, it should be able to interface with a patient safely, whereas if a work claims to be for everyday assistance, it should be portable.

\begin{figure}[h]
    \centering
    \includegraphics[width=0.45\textwidth]{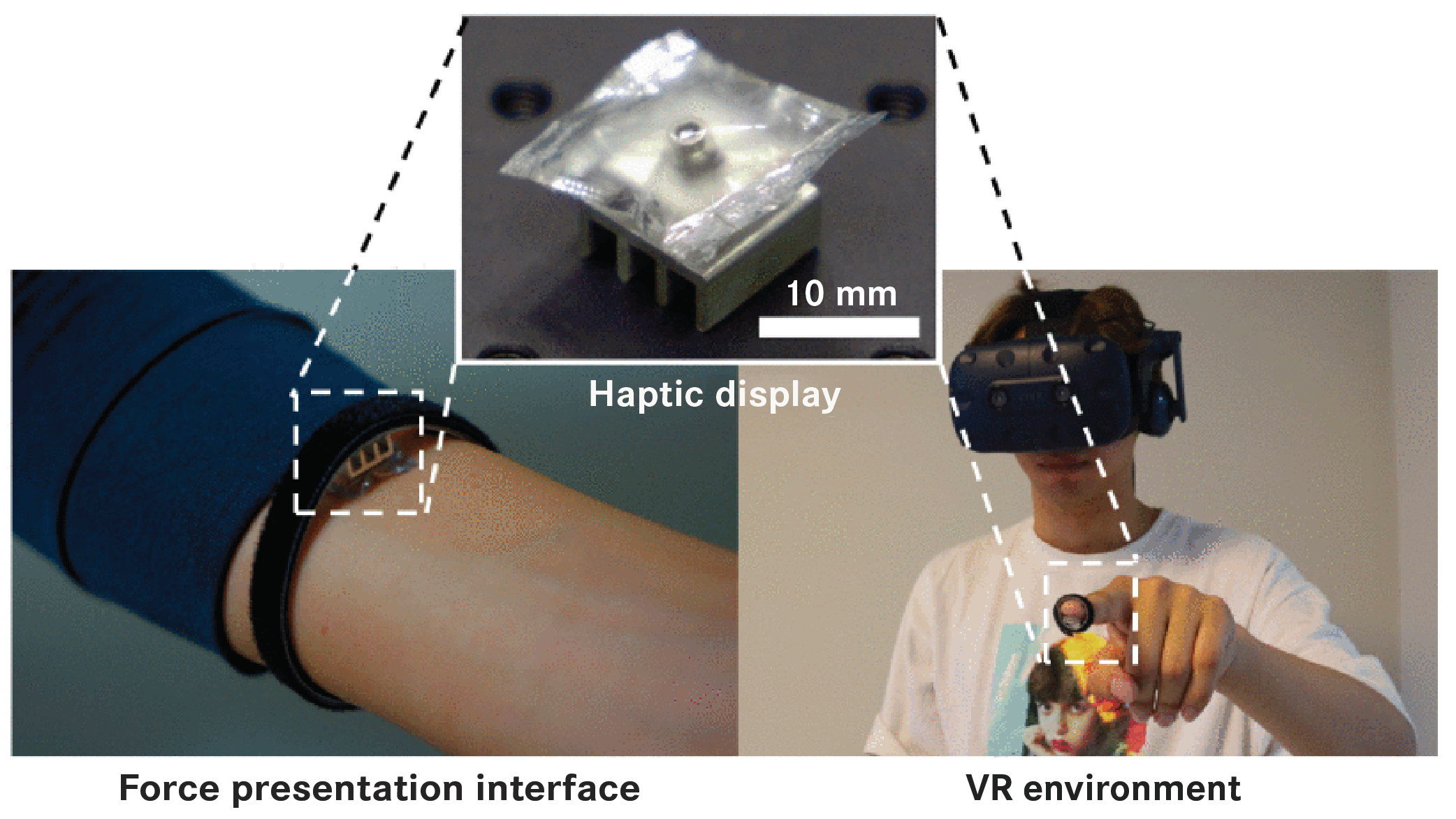}
    \caption{Haptic pouch utilizing phase-change fluid for force haptics applied to the human body, suitable for virtual reality environments.  Reproduced from Uramune et al.\cite{Uramune2022-iv} with permission.}
    \label{fig:hapouch}
\end{figure}

Virtual reality technology is a novel tool for rehabilitation in which haptics and feedback can offer useful information back to the patient as a complement or substitute to physical therapies. Virtual reality and haptics have been a popular area for applying thermo-active technology. By creating micro-arrays that are able to decoupled from a pneumatic source \cite{10052525,PUCE20196,Uramune2022-iv}, one can create tactile and visual displays that produce forces and noticeable temperature differences (Fig. \ref{fig:hapouch}).  Additionally, these haptics can help enable human-machine interfaces that give the user enhanced feedback and control \cite{https://doi.org/10.1002/aisy.202100075} \cite{https://doi.org/10.1002/adfm.202007952,https://doi.org/10.1002/adfm.202007376} Thermal management can be achieved with these technology through the addition of pumps and origami.\cite{Cacucciolo2019,act8010009}

SMPs have applications within biomedical engineering (drug delivery), aerospace (deployable, exploration components), and general fields (grippers). For example, Yuan et al.\cite{doi:10.1126/science.aaw3722} use novel materials in nanocomposite fibers to replicate artificial muscle, gaining work capacities magnitudes (60x) higher than human skeletal muscles.


Thermo-active soft actuators bring up unique applications where heat may be transferred in and out of the system from extreme ambient temperatures. Potential thermo-active actuator technologies may use principles such as photothermal \cite{JIN2022107366}, thermoelectric cooler \cite{2018JPhD...51a4005C}, and phase-change heatsinks to effectively transition between actuation and de-actuation states. Additionally, material types such as gel \cite{doi:10.34133/2019/2384347,C9TA10183G}, polyimide films\cite{Daisuke_Yamaguchi2020}, and reduced graphene oxide\cite{wangNovelReducedGraphene2019,wangElectricalActuationProperties2015,zengThermallyConductiveReduced2019} have shown capabilities for extreme temperature tolerance performances.

Some studies that aim to replicate similar functionality of traditional motors using strategies such as phase-change\cite{9044718,6907793} and SMA functionality\cite{Manfredi_2017}.  Currently, what is impeding implementation is insufficient thermal regulation for de-actuation, flexible heating/cooling pumps, and inefficient heat dissipation. However, thermo-active actuators may be most effective in extreme low-temperature environments with their current technology.

\begin{table*}[t]
\smaller
\centering
\begin{tabular}{@{}llllll@{}}
\toprule
\textbf{Material} & \textbf{Energy Density} & \textbf{Deformation} & \textbf{Bandwidth} & \textbf{Power} & \textbf{Efficiency} \\ \midrule
Human Skeletal Muscles\cite{TAYLOR2012139,doi:10.5301/jabfm.5000275} & 40 $\frac{J}{kg}$ \cite{Rich2018-hr} & 20-40$\%$ \cite{doi:10.1126/science.aaw2502} & 1-10$Hz$ \cite{Rich2018-hr} & 50-300 $\frac{kW}{m^{3}}$ \cite{Rich2018-hr} & 20$\%$ \cite{Rich2018-hr} \\ 
\hline
Fiber-based Artificial Muscles & 2-2.5$\frac{kJ}{kg}$ \cite{doi:10.1126/science.1246906} & 49$\%$ \cite{doi:10.1126/science.1246906} & 1-7.5$Hz$ \cite{doi:10.1126/science.1246906} & 27.1 $\frac{kW}{kg}$   \cite{doi:10.1126/science.1246906} & 20$\%$ \cite{doi:10.1126/science.1246906} \\
\hline
Shape Memory Polymers & 17.9$\frac{J}{g}$\cite{doi:10.1089/soro.2018.0159,10.1063/1.5033487} & 24-60$\%$ \cite{8288847,doi:10.1089/soro.2020.0080}.   & -- & 2$\frac{kJ}{kg}$\cite{doi:10.1089/soro.2018.0159,10.1063/1.5033487} & 40$\%$ \cite{https://doi.org/10.1002/adma.201300880} \\
\hline
Thermally-responsive Hydrogels & 0.35$\frac{J}{cm^3}$\cite{biomimetics3030015} & 90\% \cite{biomimetics3030015} & -- & -- & -- \\
\hline
Electrothermal Films & 10$^{-3}\frac{J}{cm^3}$ \cite{sui2011flexible} & 120\% \cite{zhou2016large} & $~$0.1Hz\cite{zhou2016large} & 0.1$\frac{W}{cm^3}$\cite{zhou2016large} & 12\%\cite{sui2011flexible} \\
\hline
Shape Memory Alloys &  $10^4$  $\frac{J}{kg}$ \cite{Rich2018-hr} & 10-50$\%$ \cite{Rich2018-hr} & 0.5-5$Hz$ \cite{Rich2018-hr} & $10^3-10^5$ $\frac{kW}{m^{3}}$\cite{Rich2018-hr} & --\\
\hline
Liquid Crystal Elastomers\cite{Guin2018,C3NR00037K,doi:10.1126/sciadv.aax5746} & 1-50$\frac{J}{kg}$ \cite{Rich2018-hr}  & 38-50$\%$\cite{He2019-jv,ford2019multifunctional} & 0.19-10Hz\cite{He2019-jv,ford2019multifunctional} & $0.80-4.5\frac{W}{cm^{3}}$\cite{Zadan2020-xb,ford2019multifunctional} & -- \\
\hline
Phase Change Actuators & 40 $\frac{kJ}{m^{3}}$ \cite{mirvakili2020actuation}  & 20-40$\%$ \cite{Tremblay:12, noguchi2020soft, mirvakili2020actuation, decroly2021optimization} & 0.15-2Hz \cite{narumi2020liquid, hiraki2020laser, Li2021-qp, Miriyev2017-ps, meder2019remotely} & $50\frac{W}{cm^{3}}$\cite{hiraki2020laser} & --  \\
\hline
\bottomrule
\end{tabular}
\caption{Properties of thermo-active actuators for desired applications, compared to human skeletal muscles. The materials provided can be used a search queries for navigating literature.}
\label{tbl:applications}
\end{table*}







\section{Current Challenges and Future Directions}
\label{discussion}
This naming scheme may seem redundant in some applications, so further discussion must be placed on the descriptive effects of a title vs keywords. 
Furthermore, the clarity of the presented examples is straightforward for both the \textbf{working principle} and \textbf{material}, but the \textbf{application/actuator type} is interesting, as one may have to merge both into a comprehensive explanation (e.g., artificial muscle $\rightarrow$ biomimetic application, contraction $\rightarrow$ actuation type).

Further discussion may be had regarding overlap between materials that are also working principles (e.g., Shape Memory Polymers) and working principles that can be materials (e.g., Phase-Change Fluids). 

The speed of thermo-active soft actuators is the most critical feature which must be improved for implementation. Currently, studies have investigated strategies for increasing bandwidth using mechanical instabilities \cite{doi:10.1089/soro.2021.0080} and incorporating thermally-conductive fillers \cite{act9030062}. Thermo-active elastomeric and phase-change actuators tend to have ineffective heating through the bulk of the device. The construction of composite materials that are effective at transferring heat faster and uniformly is necessary to increase general reliability of these actuators \cite{doi:10.1021/acsomega.8b03466,Majidi_2022,https://doi.org/10.1002/aenm.202101387}

Secondary to the bandwidth, active cooling must be implemented to further enhance the symmetry of the actuation cycles. Studies have looked to regulate the environmental temperature, or introduce heat pumps using Peltier technology \cite{Lee_2021}. This is interesting as the heating stimuli can interact with the environment, whether that be surrounding air or surfaces to simulate peristalsis \cite{C5SM02553B}.


An interesting development in soft robotics, is integrating strain-limiting layers in novel ways. Thermo-active soft actuators take this further and allow strain-limiting and functional properties to be controlled through varying the thermal conductance of a material, such as adding liquid metal into an elastomer \cite{HUANG2022100412,Usui_2021,https://doi.org/10.1002/adfm.201906098,doi:10.1021/acsami.9b19837}. Furthermore, these functional materials can be implemented into the heating stimuli to make more effecting heating junctions \cite{https://doi.org/10.1002/aenm.202201413,s19194250}.




Thermo-active soft actuators are thought to be as binary in their actuation, with the devices being ’on’ with the temperature above a critical point, or ’off’ with the temperature decreases below the same point. There has been research groups that focus on bi-stable and multi-stable approaches to pneumatic systems\cite{zou2022based} using origami could take further advantage of these stabilities using thermo-pneumatics \cite{https://doi.org/10.1002/adfm.202201891}.

More traditional origami like the Yoshiura\cite{zhang2021yoshimura} pattern and waterbomb\cite{onal2012origami} patterns have become more and more integrated into soft actuation systems \cite{Rus2018,doi:10.1073/pnas.1713450114}, as one can take advantage of natural mechanical imperfections to accomplish different motions. For thermo-active soft robots, groups have used this strategy to accomplish worm-like movement\cite{onal2012origami}, self folding\cite{doi:10.1126/sciadv.aao3865}, and crawling \cite{7519030}. Research has been done to print the actuator and structure onto paper, allowing it to be self folding and moving using joule heating\cite{7519030} (similar to Fig. \ref{fig:electrothermal}), and even having self sensing capabilities\cite{wang2018printed}.






The heating (and cooling) stimuli for these soft actuators is a necessary area of focus for these technologies to be implemented. However, the stimuli may not always be implemented into the naming scheme, as some studies do not have this as a focus (e.g., use a climate chamber, arbitrary heating, etc.). Therefore, it is recommended that this aspect is reflected in the novelty of the work to increase the impactfulness. Sogabe et al. \cite{SOGABE2023114587} demonstrate the ability of a temperature-controlled water tank in dramatically increasing the bandwidth of hollow elastomers filled with phase-change fluids. Furthermore, we can conclude that the key factor in the clarity of a name is the balance of simplicity and descriptiveness, while the key factor for impactfulness is broken down to less of a science, consisting of memorability and evocativeness. Studies such as Zadan et al. \cite{https://doi.org/10.1002/adma.202200857} investigate the ability of thermo-electric devices (TEDs) for reversible heating and cooling. These devices tend to overheat and require additional external cooling for enhanced efficiency. Thus, underwater environments have been investigated which improved cooling capabilities of the TEDs.

Another vital discussion is on the importance of marketability and potential branding. For example, a popular example within dielectric elastomer actuators (DEAs) is the HASEL (Hydraulically Amplified Self-healing ELectrostatic) actuator. Having an acronym for your naming scheme is not necessary, but may allow for additional iterations that can add information without becoming too wordy (e.g., Peano-HASEL \cite{doi:10.1126/scirobotics.aar3276} following HASEL \cite{doi:10.1126/science.aao6139}). This allows researchers to introduce novel aspects about their device and carry necessary information while staying memorable. In following the proposed naming scheme, it can lead to a title becoming too wordy which can have a detrimental effect on impact (number of citations) \cite{RePEc:spr:scient:v:76:y:2008:i:1:d:10.1007_s11192-007-1892-8}.

As thermo-active soft actuators are improved, the unique functionality of these kinds of devices will add to the draw of the technology. For example, self-healing capabilities of materials using Diels Alder chemistry \cite{act9020034,TERRYN2021187} have been investigated for pneumatic actuators \cite{Wang2012-va}. With further research into thermo-pneumatics \cite{Yoon2023-jy,doi:10.1089/soro.2022.0170}, these technologies may be combined for enhanced properties. Self-sensing \cite{9140392} is another area of focus, as the more function and 'intelligence' that can be consolidated to the material and geometry, the less control system considerations are required.

Generalization of the naming scheme for maximum clarity is beneficial for enhancing the visibility of ones paper, but it is still beneficial to make papers stand out by clearly defining novel features within a work. For example, the concept of 'thermo-pneumatics' has been described in different ways before getting defined. Mirvakili \cite{mirvakili2020actuation} defines this concept as 'untethered pneumatic artificial muscles' within the title, while a derivative study by Yoon \cite{Yoon2023-jy} uses 'untethered [and] pumpless phase change soft actuators' in the title, but goes onto describing a concept of thermo-pneumatics where heating is the stimuli that causes this 'pneumatic' volume expansion. However, 'pumpless' still describes the novel aspect with less clarity, which brings up the question of whether novel is more appealing to the reader than clarity. Hence, clear descriptions should be included in the keywords section of a submission to allow for more transparency in searching for articles. 

Ten studies were selected to use for Table \ref{tbl:naming}, and highlighted with colors that correspond to the clarity of the title. These colors are combinations of the three-part naming scheme with working principle, material, and application. Titles of the studies have been attenuated for space requirements. The majority of the papers satisfied $\frac{2}{3}$ of the naming scheme, although some characteristics such as applications were kept general (e.g., 'soft actuators', 'soft robotics'). The key to ensuring clarity is to verify the naming scheme is met either in the title or that relevant keywords have been defined by authors.

The role of soft actuators is constantly being redefined, and each subtype is finding its own niche in applications. Further, thermo-active soft actuators must be fully evaluated for applications where they are standalone, and applications where they synergize and cooperate with traditional systems (e.g., end-effector, thermal management, quick release). 
The modeling strategies for expansive soft actuators and robotics are not fully developed yet, but heat transfer simulation has been widely used and understood, leading to possibilities of multi-domain modeling in flexible but not stretchable systems.  

Although it was out of the scope of this review, machine learning integration to the article scraping could be beneficial for achieving more optimal titles.


\section{Conclusion}\label{Conclusion}
Deciding a name, title, and keywords for your technology in an academic setting may be an afterthought, but dedicating time to simplifying your idea using descriptive terms can enhance the discoverability of your work and further clarify ones scope. Hence, the overarching takeaway from this review is this: \textit{to maximize the findability of ones study, you must optimize the clarity of the title and accompanying keywords}. We recommend using a combinatory title where a proposed name or novel feature is introduced followed by a clear overview of material, working principle, and application. Furthermore, thermo-active soft actuators offer a unique challenge where one must identify the various domains the technology overlaps and apply naming appropriately to ensure clarity. Future studies with thermo-active soft actuators should fully realize enhanced controllability, bandwidth, and thermal regulation for increasing overall efficiency. With thermo-active applications spanning a plethora of applications dealing with external and internal physical Human-Robot Interactions, further clarification can enable researchers to pick the best tool for an application. Where phase-change materials can be implemented to create thermo-pneumatic systems or stiffness control for artificial muscles, shape memory materials can be used to accomplish delicate grasping tasks, and nanostructured fiber-based systems can accomplish high-bandwidth goals. With such a wide range of applications, each demanding specialized focus and terminology, naming becomes a crucial step for enhancing the visibility and impact of your work across the diverse landscape of soft robotics research.

\section*{ACKNOWLEDGMENT}

This work was funded by National Science Foundation NSF under grant number 2045177, and by the National Institute of Health (NIH) through grant T32GM136501.

\clearpage
\begin{sidewaystable}
\vspace{8cm}
\newcolumntype{Y}{>{\centering\arraybackslash}X}
\begin{tabularx}{\linewidth}{|Y|Y|Y|Y|Y|Y|Y|Y|}
\hline
\textbf{Study} & \textbf{Material} & \textbf{Actuation Movement} & \textbf{Actuation Stimuli} & \textbf{Application} & \textbf{Novel Aspect} & \textbf{Journal} & \textbf{Impact Factor (2 year)} \\
\hline
\rowcolor{red!25} Soft Material for Soft Actuators\cite{Miriyev2017-ps}& Phase-Change Fluid, Silicone Elastomer & Expansion, Contraction & Joule Heating & Artificial Muscle & Working Principle (Phase-Change) & Nature Communications & 16.6 \\
\hline
\rowcolor{orange!25} NiTi Soft Robotics\cite{5354178} & Nitinol & Contraction & Shape Memory & Artificial Muscle Fiber & Micro & IROS & 2.45 \\
\hline
\rowcolor{orange!25} Block Copolymer Muscles\cite{Lang2022-jy} & ABA Triblock Copolymer & Contraction, Rotation & Flat Iron, Hydration & Artificial Muscle Fiber & Hydro and Heat Activated (Micro) & Nature Nanotechnology & 38.3 \\
\hline
\rowcolor{purple!25} Thermo-Activated Micro-Actuator\cite{PUCE20196} & Galinstan & Expansion & Joule Heating & Tactile Display & Micro & Microelectronic Engineering & 2.3 \\
\hline
\rowcolor{purple!25} HaPouch Display\cite{Uramune2022-iv} & Nylon Film, Phase-Change Fluid & Inflation & Peltier & Haptic Display & Haptics & IEEE Access & 3.9 \\
\hline
\rowcolor{black!25} Tri-Layer Electro-Thermal Actuators\cite{doi:10.1089/soro.2018.0159} & Carbon Nanotubes, Kapton, polycaprolacton & Bending & Joule Heating & Soft Robots & Out-of-Plane Stuctures & Soft Robotics & 7.9 \\
\hline
\rowcolor{purple!25} Magnetic Pneumatic Muscles\cite{mirvakili2020actuation} & Balloon, Phase-Change Fluid & Expansion & Induction & Artificial Muscle & Untethered Pneumatics & Science Robotics & 25.0 \\
\hline
\rowcolor{purple!25} Bioinspired Phase Change Actuators\cite{Yoon2023-jy} & Silicone & Expansion & Peltier & Soft Gripper, Soft Robots & Untethered, Pumpless & Chemical Engineering Journal & 15.1 \\
\hline
\rowcolor{orange!25} LC Elastomer\cite{https://doi.org/10.1002/adma.202200857} & Liquid Crystal Elastomer & Bending & Peltier & Soft Robots & Energy Harvesting, Active Cooling & Advanced Materials & 29.4 \\
\hline
\rowcolor{orange!25} Fishing Line Muscles\cite{doi:10.1126/science.1246906} & Fishing Line, Sewing Thread & Contraction & Joule Heating & Artificial Muscles & -- & Science & 56.9 \\
\hline
\hline
\end{tabularx}

\begin{tabular}{ll}
\textcolor{blue!40}{\rule{1cm}{1.5mm}} & Working Principle \\
\textcolor{yellow!40}{\rule{1cm}{1.5mm}} & Material \\
\textcolor{red!40}{\rule{1cm}{1.5mm}} & Application \\
\color{green!40}{\rule{1cm}{1.5mm}} & Working Principle + Material \\
\color{purple!40}{\rule{1cm}{1.5mm}} & Working Principle + Application \\
\color{orange!40}{\rule{1cm}{1.5mm}} & Material + Application \\
\color{black!40}{\rule{1cm}{1.5mm}} & Working Principle + Material + Application \\
\end{tabular}
\caption{Comparing the naming schemes of select current studies. Rows are highlighted with what attribute is clearly stated in the title. Color key is shown for all combinations, even if not all are represented in the table.}
\label{tbl:naming}
\end{sidewaystable}

\clearpage


\bibliography{biblio}

\end{document}